\documentclass[sigconf]{acmart}
\usepackage[english]{babel}
\usepackage{blindtext}

\usepackage{etoolbox}

\usepackage{lipsum}

\makeatletter
\patchcmd{\maketitle}{\@copyrightspace}{}{}{}
\makeatother

\usepackage{siunitx}
\usepackage{pifont}
\usepackage{graphicx}
\usepackage{amsmath, amsthm, amssymb,textcomp}
\usepackage{algorithm} 
\usepackage{booktabs}
\usepackage{url}
\usepackage{subcaption} 
\usepackage{algpseudocode}
\usepackage{paralist}
\usepackage{color}
\usepackage{relsize}
\usepackage{stmaryrd}
\usepackage{epsfig} 
\usepackage{bm} 
\usepackage{listings}
\usepackage{color}
\usepackage{mathtools}
\usepackage{mathrsfs}
\usepackage{xspace}
\usepackage{soul,xcolor}
\usepackage{float}
\usepackage{verbatim}
\usepackage{epstopdf}

\usepackage[utf8]{inputenc}
\usepackage{calrsfs}

\DeclareMathOperator*{\argminB}{argmin}   

\newcommand{\sys}{D-SLATS} 
\newcommand{\algone}{DKAL} 
\newcommand{\algtwo}{DKALarge} 
\newcommand{\algthree}{DOPT} 

\begin{document}
%
\title{D-SLATS: Distributed Simultaneous Localization and Time Synchronization}
\author{Amr Alanwar}
\affiliation{University of California, Los Angeles}
\email{alanwar@ucla.edu}

\author{Henrique Ferraz}
\affiliation{University of California, Santa Barbara}
\email{henrique@ece.ucsb.edu}

\author{Kevin Hsieh}
\affiliation{University of California, Los Angeles}
\email{khsieh39@ucla.edu}

\author{Rohit Thazhath}
\affiliation{University of California, Los Angeles}
\email{rohitbhasi@ucla.edu}

\author{Paul Martin}
\affiliation{University of California, Los Angeles}
\email{pdmartin@ucla.edu}

\author{Jo\~{a}o Hespanha}
\affiliation{University of California, Santa Barbara}
\email{hespanha@ece.ucsb.edu}

\author{Mani Srivastava}
\affiliation{University of California, Los Angeles}
\email{mbs@ucla.edu}

\renewcommand{\shortauthors}{A. Alanwar et al.}

\copyrightyear{2017}
\acmYear{2017}
\setcopyright{acmcopyright}
\acmConference{Mobihoc '17}{July 10-14, 2017}{Chennai, India}\acmPrice{15.00}\acmDOI{http://dx.doi.org/10.1145/3084041.3084049}
\acmISBN{978-1-4503-4912-3/17/07}

\begin{abstract}
Through the last decade, we have witnessed a surge of Internet of Things (IoT) devices, and with that a greater need to choreograph their actions across both time and space. Although these two problems, namely time synchronization and localization, share many aspects in common, they are traditionally treated separately or combined on centralized approaches that results in an inefficient use of resources, or in solutions that are not scalable in terms of the number of IoT devices. Therefore, we propose \sys{}, a framework comprised of three different and independent algorithms to jointly solve time synchronization and localization problems in a distributed fashion. The first two algorithms are based mainly on the distributed Extended Kalman Filter (EKF) whereas the third one uses optimization techniques. No fusion center is required, and the devices only communicate with their neighbors. The proposed methods are evaluated on custom Ultra-Wideband communication Testbed and a quadrotor, representing a network of both static and mobile nodes. Our algorithms achieve up to three microseconds time synchronization accuracy and 30 cm localization error.

\end{abstract}


\maketitle
\section{Introduction}\label{sec:intro}

With the dramatic increase in the number of wireless devices, it is important to coordinate timing among networked devices and to provide contextual information, such as location.  Maintaining a shared notion of time is critical to the performance of many cyber-physical systems (CPS) such as wireless sensor networks, Big Science \cite{conf:rabbit}, swarm robotics \cite{conf:formationctrl}, high frequency trading \cite{conf:ptp}, telesurgery \cite{conf:surgnet}, and global-scale databases \cite{conf:Gspanner}. In addition, position estimation is necessary for different fields such as military \cite{conf:military}, indoor and outdoor localization \cite{conf:indoor}.

Since time and distance are causally associated with each other, seeking a solution for one of them provides information about the other. In many cases, accurate time synchronization among nodes is necessary for their precise localization, as is the case in techniques based on time-of-arrival (TOA) \cite{conf:toa} and time-difference-of-arrival (TDOA) \cite{conf:tdoa} measurements. Moreover, a combined  solution to time synchronization and localization problems in sensor networks could be obtained with less computational effort by tackling the two problems in a unified approach, instead of considering them as two separate problems. This type of approach has been addressed in previous works. The effect of clock mismatch on the accuracy of ranging systems is studied in \cite{uwb_toa_clock_bias}, where TOA measurements are considered, and in \cite{pos_unknown_skew}, where a maximum likelihood estimator for TDOA-based positioning was proposed. In \cite{conf:robust} localization is investigated based on time of flight measurements for asynchronous networks using least squares methods. Least squares are also used in \cite{conf:joint} to achieve synchronization using TOA measurements, and in \cite{slats_unknown_node_wsn} with accurate and inaccurate network anchors. In \cite{uwb_toa_distributed}, the authors derived a distributed maximum log-likelihood estimator that fuses clock offset, bias, and range estimates. While these approaches present satisfactory theoretical results, they either make several non-practical assumptions of the underlying network, such as perfect synchronization among anchor nodes, or they lack experimental hardware evaluation of the proposed methods.

Centralized algorithms for localization and time synchronization, though theoretically might attain the optimal solution, are neither robust nor scalable to complex large-scale dynamical systems, where the nodes are distributed over a large geographical region. In order to perform simultaneous time synchronization and localization using centralized algorithms, all nodes must send their measurements to a fusion center that computes the estimates of position and clock parameters for every node. The information is then sent back to every node and the process is repeated. This strategy requires a large communication overhead, might not be energy efficient, and has a potentially critical failure point at the fusion center.



This paper introduces \sys{}, a framework that is comprised of three different, independent, and distributed algorithms to achieve network time synchronization and accurate position estimates using time stamped message exchanges, filters and optimization techniques. \algone{} is the first proposed algorithm which is based mainly on the distributed Extended Kalman Filter (EKF) with diffusion between wireless nodes. \algone{} improves some aspects of \cite{conf:disali} by decreasing some of the computational efforts. The second algorithm, \algtwo{} is based on the results from \cite{conf:dislarge} and deals with extremely large scale systems where \algone{} might not be the best option. It is important to note that the modifications proposed in our work to the algorithms presented in \cite{conf:dislarge} and \cite{conf:disali}, can be used in other domains as well. Finally, we present \algthree{}, an optimization technique to synchronize and localize the wireless nodes in a distributed fashion. Since \sys{} operates in a distributed fashion, it does not require a fusion center and the results from the three algorithms can work perfectly with different network topologies.

The three algorithms proposed are evaluated on custom ultra-wideband wireless hardware for networks with different topologies containing both static and mobile nodes. This work leverages ultra-wideband RF communication to make precise timing measurements. While UWB has improved non-line-of-sight performance in comparison to signal strength methods like those based on RSSI, it should be noted that UWB timing accuracy can deteriorate with increased environmental clutter and signal attenuation, as described in \cite{jan}. We compare the proposed algorithms with a conventional centralized EKF, and we present the accuracy of the node position estimation, as well as, the time synchronization.

The rest of the paper is organized as the following: Related work is shown in Section \ref{sec:relwrk}. Section \ref{sec:system} provides an overall overview about the system model. We then go through \sys{} algorithm by algorithm in Section \ref{sec:alg}. Then, Section \ref{sec:eval} evaluates the introduced algorithms on static and mobile network of nodes. Finally, Section \ref{sec:conc} concludes this paper. 
 




\section{Related Work} \label{sec:relwrk}

We investigate the related work in concurrent synchronization and localization. One category of previous work  was based on distributed maximum likelihood estimators, theoretical Cram\'{e}r Rao Lower Bounds (CRLB), batch estimation techniques such as least squares regression, and experimental heuristics. Gholami \emph{et al.} \cite{pos_unknown_skew} derived a maximum likelihood estimator (MLE) for joint clock skew and position estimation. However, they assume that a number of reference nodes are perfectly synchronized with a reference clock and transmit their signals at a common time instant. Similarly, joint time synchronization and localization is solved in \cite{slats_unknown_node_wsn} with accurate and  inaccurate anchors in terms of time synchronization and localization. In \cite{uwb_toa_distributed}, Denis \emph{et al.} derived a distributed maximum log likelihood estimator that fuses clock offset, bias, and range estimates. While these approaches outline theoretical estimators, they make several non practical assumptions of the underlying network such as perfect synchronization among anchor nodes, and do not evaluate their methods on actual hardware. Recently, ATLAS estimates locations using a maximum likelihood method and known-position beacon transmitters to synchronize the clocks of receivers \cite{conf:atlas}. Positioning using time-difference of arrival measurements is proposed in \cite{tdoa}.

Lately, accurate RF time-stamping, spearheaded by impulse-radio ultra-wideband (IR-UWB) devices \cite{uwb_ranging}, has encouraged research on low error ranging techniques and their related time synchronization. Polypoint \cite{polypoint}, for instance, introduced a RF localization system which enables the real-time tracking and navigating of quadrotors through complex indoor environments. Another quadrotor localization framework was introduced in \cite{decawave_quad_ipin}. Furthermore, clock bias measurements were used in \cite{uwb_quad_oneway} to allow for simpler range measurements using one-way TOA/TDOA messages compared to the  more expensive two-way protocols. These works are based on the popular DecaWave DW1000 radio \cite{dw1000}. Recent developments in UWB communications offer high precision positioning through which a new range of applications is enabled \cite{conf:consolu}. Moreover, \cite{polypoint} demonstrated a 56 cm localization error with frequency 20 Hz and density 0.005 node/$m^3$. On the other hand, \cite{uwb_quad_oneway} achieves a 28 cm localization error. However, it used 100 Hz frequency with 0.037 node/$m^3$. Finally, \cite{decawave_quad_ipin} shows a 20 cm localization error with frequency 10 Hz given a density of 0.054 node/$m^3$. 


 A linear data model for joint localization and clock synchronization is proposed in \cite{joint_loc_sync_wsn}. Its estimates included batch offline least squares methods. Estimators to precisely estimate range and clock parameters
from the measurements are suggested in \cite{joint_range_clock_param}. It used master-slave clock synchronization to improve round-trip time estimates. Impact of clock frequency offsets on the accuracy of ranging systems based on TOA measurements is analyzed in \cite{uwb_toa_clock_bias}. More application-specific solutions include using unmanned aerial vehicles to relay GPS information to a network of energy constrained nodes, who then localize and synchronize themselves to the mobile GPS receiver, saving energy \cite{joint_loc_uav} and leveraging clock bias estimation to improve RF time-of-flight estimates for non-UWB transceivers \cite{cheap_nonuwb_slats}. For more complete treatments of localization and synchronization methods, we refer the reader to \cite{patwari_coop} and \cite{sync_survey}, respectively.


\vspace*{0.15in}
\section{System Model}
\label{sec:system}

Consider a set of $N$ nodes indexed by $k\in \{0, \ldots, N-1\}$ distributed geographically over some region. We say that two nodes are connected if they can communicate with each other, and we denote the neighborhood of a given node by the set $\mathcal{N}_k$ that contains all the nodes that are connected to node $k$. Our state-space model is 
\begin{equation}
\begin{split}
x^{k}_{i+1} &= f(x^{k}_i) + n^k_i\\
y^{k,j}_i &= h^j(x^k_i) + v^k_i,\\
\end{split}
\end{equation}
where the state of node $k$ at time $i$ is denoted by $x^k_i$, and the measurement available to node $k$ from the neighborhood node $j \in \mathcal{N}_k$ is represented by $y^{k,j}_i$. The process and measurement noise are $n^k_i$ and $v^k_i$, respectively, and assumed to be Gaussian. The state update function is $f$ and $h$ is the measurement function. The state vector consists of three components, $x^k_i =\left[{\bm{p}^k_i}^T,\: o^k_i,\: b^k_i\right]^{T}$, where $\bm{p}^k_i$ is the three dimensional position vector, $o^k_i$ is the clock time offset, and $b^k_i$ is the clock frequency bias. We adopt a convention where both  $o^k_i$ and  $b^k_i$ are described with respect to the master node, which can be any node. It is recommended to pick the master node in the middle of the network graph to achieve faster time synchronization. It is assumed that the nodes are static and that the clock parameters evolve according to the first-order affine approximation of the following dynamics,

\begin{equation}\label{eq:clockModel}
\begin{aligned}
&o^k_{i+1} = o^k_{i} + b^k_{i} \delta_t \\
&b^k_{i+1} = b^k_i
\end{aligned}
\end{equation}
where $\delta_t:=t_M(i+1)-t_M(i)$ given that $t_M$ is the root node time which is the global time. Therefore, we can write the update function for a static node as

\begin{eqnarray}
f( x^k_i) &= &\begin{bmatrix}
\bm{p}^k_i\\
o^k_i+b^k_i\delta_{t}\\
b^k_i
\end{bmatrix}
\end{eqnarray}



\subsection{Measurement types}

\begin{figure*}
\centering{}\includegraphics[width=1\textwidth]{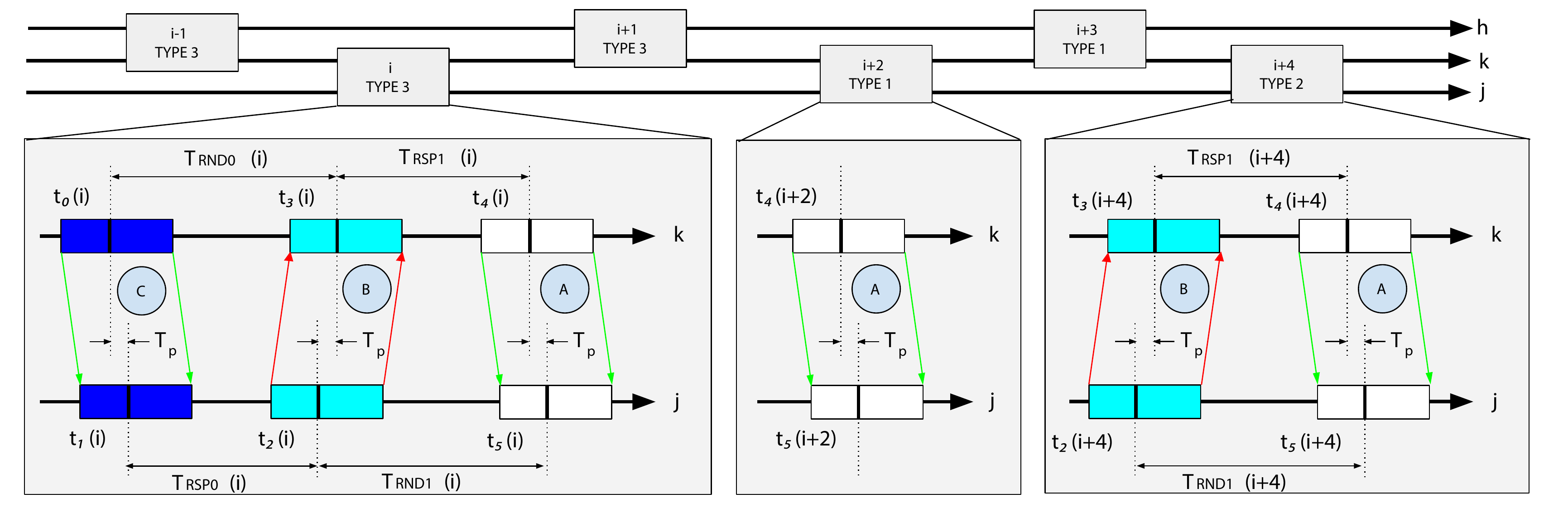}\protect\caption{The top of this diagram shows six synchronization events between three devices, labeled $h$, $k$, and $j$. Each event is  classified as TYPE 1, TYPE 2 or TYPE 3 depending on the number of transmissions sent.}
\label{fig:msg_types2}
\end{figure*}

The \sys{} architecture supports three types of measurements which are distinguished by the number of messages exchanged between a pair of nodes. These measurements types are shown in detail in Figure \ref{fig:msg_types2}, where time stamps $t_0(i)$ through $t_5(i)$ denote the locally measured transmission (TX) and reception (RX) times stamps, and $T_{RSP}(i)$ and $T_{RND}(i)$ define, respectively, the response and the round-trip durations between the appropriate pair of these timestamps. The propagation velocity of radio is taken to be the speed of light in a vacuum, denoted  by $c$.

The three message types are:

\begin{itemize}
\item \textbf{TYPE 1}: a single transmission is sent from $j$ to $k$, yielding two timestamps, one from the TX instant and another from the RX instant. TYPE 1 results in finding the \textbf{counter difference} at time $i$ denoted by $d^{k,j}_i$ which is the measurement of the difference between the clocks of each node. This is a \textbf{time measurement} that includes the effects of propagation delay $T_p$.
\begin{align}
d^{k,j}_i &= t_{5}(i)-t_{4}(i)
\end{align}

\item \textbf{TYPE 2}: when a TYPE 1 message is followed by a reply message we allow for round-trip timing calculations. TYPE 2 results in the \textbf{single-sided two-way range} $r^{k,j}_i$ which is a \textbf{distance measurement} between pair of nodes j and k, with error proportional to the response turnaround time $T_{RSP1}$. This is a noisy measurement due to frequency bias discrepancies between $k$ and $j$. 
\begin{align}
r^{k,j}_i &= \frac{c}{2}\left(T_{RND1}(i)-T_{RSP1}(i)\right)
\end{align}
\item \textbf{TYPE 3}: one last transmission completes a handshake trio allowing for a more precise round-trip timing calculation. Type 3 results in finding the \textbf{double-sided two-way range} $R^{k,j}_i$ which is another \textbf{distance measurement} between nodes $k$ and $j$ based on a trio of messages between the nodes at time $i$. The error is proportional to the relative frequency bias between the two devices integrated over the period. This is a more accurate estimate than $r^{k,j}_i$ due to mitigation of frequency bias errors from the additional message. 
\begin{align}
R^{k,j}_i &=c\dfrac{T_{RND0}(i)T_{RND1}(i) - T_{RSP0}(i)T_{RSP1}(i)}{T_{RND0}(i) + T_{RND1}(i) + T_{RSP0}(i) + T_{RSP1}(i)}
\end{align}
\end{itemize}


With that, the measurement vector at node $k$, from node $j \in \mathcal{N}_k$ has the form
\begin{eqnarray}
y^{k,j}_i &= &\begin{bmatrix}
d^{k,j}_i\\
r^{k,j}_i\\
R^{k,j}_i
\end{bmatrix}
\end{eqnarray}


It is important to note that  a subset of these measurements may be used rather than the full set, i.e, we can have experiments involving just $r^{k,j}_i$, $R^{k,j}_i$, or $d^{k,j}_i$. The three measurements types can be translated to the state vector according to the clock model from $\eqref{eq:clockModel}$ and the physics relating the propagation speed, the duration between the propagation and the distance between the nodes. Thus, we have measurement function in following form

\begin{align}
h^j(x^k_i)&=\left[\begin{array}{c}
\left(o^j_i-o^k_i\right) + \left\Vert \bm{p}^{j}_i-\bm{p}^{k}_i\right\Vert _{2}/c\\
\left\Vert \bm{p}^{j}_i-\bm{p}^{k}_i\right\Vert _{2}+\frac{c}{2}\left(b^{j}_i-b^{k}_i\right)T_{RSP1}\\
\left\Vert \bm{p}^{j}_i-\bm{p}^k_{i}\right\Vert _{2} + c\cdot \tilde{R}^{k,j}_i\\
\end{array}\right]
\label{eq:measmodel}
\end{align}
where
{\small \begin{align*}
&\tilde{R}^{k,j}_i := \\ &\dfrac{(b^k_i-b^j_i)(T_{RND0}(i)T_{RND1}(i) - T_{RSP0}(i)T_{RSP1}(i))}{ (1+ b^k_i-b^j_i)T_{RND0}(i) + T_{RND1}(i) + T_{RSP0}(i) + ( 1 + b^k_i-b^j_i)T_{RPS1}(i)}
\end{align*}}

\section{Algorithms} \label{sec:alg}
The time synchronization and localization problem focus on finding an estimate for the clock parameters and the 3D position (i.e the state of the system) using the timestamp measurements and the model of the system. We denote by $\hat{x}^k_{i|l}$ the estimate of $x^k_i$ given the observations up to time $l$ where every node seeks to minimize the mean-square error $\text{E} \| x^k_i-\hat{x}^k_{i|l}\|^2$.

\subsection{\algone}\label{sec:alg1}
Our \algone{} algorithm is derived from the Diffusion Extended Kalman filter presented in \cite{conf:disali}, which uses the information form rather than the conventional form. The advantage of the information form over the covariance form becomes more evident for some categories of problems. For instance, the information filter form is easier to distribute, initialize, and fuse compared to the covariance filter form in multisensor environments \cite{conf:ACC}. Furthermore, it can reduce dramatically the computation and storage which is involved in the estimation of specific classes of large-scale interconnected systems \cite{conf:app1977}. Also, the update equations for the estimator are computationally simpler than the equations of the covariance form. 

By proposing \algone{}, we are seeking a distributed implementation that avoids the use of a fusion center and instead distributes the processing and communication across the sensor network. Among distributed algorithms, diffusion algorithms are amenable for easy real-time implementations, for good performance in terms of synchronization and localization accuracy, and for the fact that they are robust to node and link failure. Some of the disadvantages of the algorithm in \cite{conf:disali} that we try to address include:
\vspace{-0.5mm}
\begin{itemize}
\item Taking the inverse of a large matrix is not feasible on embedded devices. A large-scale network implies inverting an error covariance matrix $P^k_{i|i}$ on each node with dimension $N \times N$.
\item Every node monitors the whole network state which is a disadvantage in large scale networks, as the size of $x$ and $P$ will increase dramatically.
\end{itemize}
\vspace{-0.5mm}
To address the first issue, we propose modifying the measurement update in Algorithm 2 in \cite{conf:disali} by using the Binomial inverse theorem. This provides a formula to compute $P^k_{i|i}$ from $P^k_{i|i-1}$ without having to invert any $N \times N$ matrix. We propose the following modifications:

\vspace{-3mm}
\begin{align}
\tilde{Q}_0^{-1} &:= P_{i|i-1}^{k^{-1}}\\
 \tilde{Q}_{j+1}^{-1} &:= \tilde{Q}_{j}^{-1} + U_{j}B_{j}V_{j} \quad \quad \forall j \in {1,2,...,n}
\end{align}
where $U_j = \hat{H}_{i}^{{*}^{k,j}}$, $B_j = {R_{i}^{j}}^{-1} $ and $V_j = \hat{H}^{k,j}_{i}$. Thus we have $\tilde{Q}_N^{-1} := P_{i|i}^{{-1}^k}$. Since the Binomial inverse theorem states that
\begin{align}
\left({ {A}}+{ {UBV}}\right)^{{-1}}={ {A}}^{{-1}}-{ {A}}^{{-1}}{ {U}}\left({ {B}}^{{-1}}+{ {VA}}^{{-1}}{ {U}}\right)^{{-1}}{ {VA}}^{{-1}},
\end{align}
It follows that
\begin{align}
\tilde{Q}_{j+1} &= (\tilde{Q}_j^{-1} + U_jB_jV_j )^{-1}\\
		&= \tilde{Q}_j - \tilde{Q}_jU_j(B^{-1} + V_j \tilde{Q}_j U_j)^{-1}V\tilde{Q}_j.
\end{align}

The other issue of monitoring the whole network state will be addressed by the \algtwo{} and \algthree{} algorithms. The DKAL algorithm is comprised of three main steps: measurement update, diffusion update section, and time update steps. First, we define the following matrices obtained by linearizing the state update and the measurement update functions around some point $\psi^k$.

\begin{eqnarray}
&\bar{F}(\psi^k) := \frac{\partial f(x^k)}{\partial x^k}\Big\rvert_{x^k=\psi^k} \qquad \bar{H}^j(\psi^k) := \frac{\partial h^j(x^k)}{\partial x^k}\Big\rvert_{x^k=\psi^k}\\
&\bar{u}^k_{i}(\psi^k) :=  f(\psi^k) - \bar{F}(\psi^k)\psi^k. 
\end{eqnarray}

The algorithm starts with the measurement update where every node $k$ obtains $\psi ^k_{i}$ at time step $i$. Next, in the diffusion step, information from the neighbors of node $k$ are combined in a convex manner to produce a new state estimate for the node. Finally, every node performs the time update step. The proposed algorithm can be summarized as shown in Algorithm 1.

\begin{algorithm}[ht]
\label{alg:two}
\begin{flushleft}
\textbf{Algorithm 1: DKAL}\\
Start with $\hat{x}^k_{0|-1}=\mathbf{E}x_0$ and $P^k_{0|-1}= \Pi_0$ for all $k$, and at every time instant $i$, compute at every node $k$:\\
\textbf{Step 1}: Measurement update:
\begin{eqnarray}
\hat{H}^{k,j}_i&=& \bar{H}^j(\hat{x}^k_{i|i-1})\label{eqn:1}\\
\tilde{Q}_0 &=& P_{i|i-1}^{k}\\
\tilde{Q}_{j+1}  &=& \tilde{Q}_j - \tilde{Q}_j\hat{H}_{i}^{*^{k,j}}(R_{i}^{^j} + \hat{H}^{k,j}_{i} \tilde{Q}_j \hat{H}_{i}^{*^{k,j}})^{-1}\hat{H}^{k,l}_{i} \tilde{Q}_j \quad \forall j \in \mathcal{N}_{k}\nonumber\\
P_{i|i}^{k} &=& \tilde{Q}_N \\
\psi ^k_{i}&=& \hat{x}^k_{i|i-1} + P^k_{i|i}\sum\limits_{j\in\mathcal{N}_{k}}\hat{H}^{*^{k,j}}_{i}R_{i}^{{-1}^j}[y^{k,j}_{i}-h^j(\hat{x}^k_{i|i-1})]\nonumber\\
\end{eqnarray}
\textbf{Step 2}: Diffusion update:\\
\begin{eqnarray}
[\hat{x}^k_{i|i}]_{m}&\leftarrow &\sum\limits_{j\in\mathcal{N}_{k}}c^{k,j}_{m}[\psi^{j}_{i}]_{m}
\end{eqnarray}
\textbf{Step 3}: Time update:
\begin{eqnarray}
\hat{x}^k_{i+1|i} &= &\bar{F}_{i}(\hat{x}^k_{i|i}) \hat{x}^k_{i|i} + \bar{u}^k_{i}(\hat{x}^k_{i|i})\\
P^k_{i+1|i} &= &\bar{F}_{i}(\hat{x}^k_{i|i})P^k_{i|i}\bar{F}_{i}(\hat{x}^k_{i|i})^{*} + Q_{i}
\end{eqnarray}
\end{flushleft}
\end{algorithm}
In the algorithm, $Q_i$ and $R^j_i$ are the process covariance matrix at time $i$, and the measurement noise covariance matrix of node $j$ at time $i$, respectively. The $c^{k,j}_m$ elements represent the weights that are used by the diffusion algorithm to combine neighborhood estimates.




\subsection{\algtwo} \label{sec:algo2}

\begin{figure}[tb]
\centering
\includegraphics*[width = 0.35\textwidth]{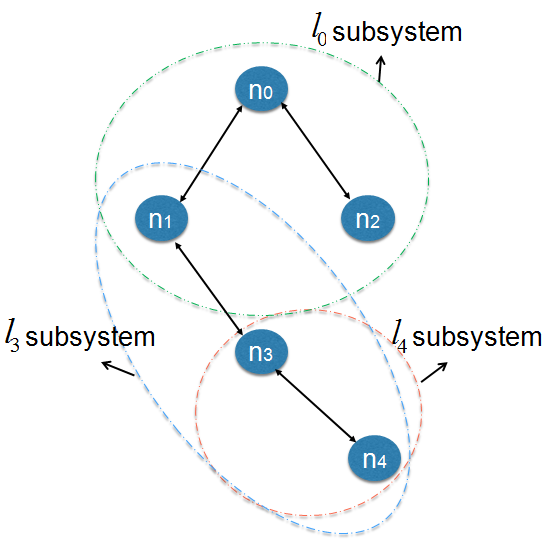}
\caption{An example of how a system can be divided in three subsystems.}
\label{fig:subsys_ex}
\end{figure}

\begin{figure}[tb]
\centering
\includegraphics[width = 0.5\textwidth]{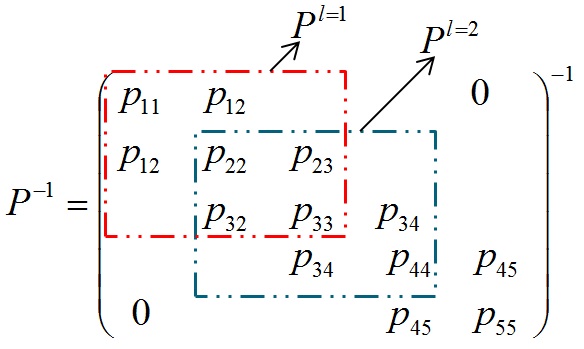}
\caption{Example of $P^{(l)}$ values out of original $P$.}
\label{fig:Pl}
\end{figure}

As mentioned before, the disadvantage of \algone{} relies on the fact that  each node in the network has to monitor the state of the entire network and carried out expensive matrix operations. Therefore, we are going to tackle these disadvantages by proposing \algtwo{}. 

The key idea behind this algorithm is to let every node monitor only its neighbors, or we can say its subsystem. Therefore, the size of the $x$ and $P$ will depend only on the number of neighbors which solves the two disadvantages of \algone{}. \algtwo{} spatially decomposes a sparsely connected large-scale network before carrying out Kalman filtering operations. Network connectivity is used to break down the global system into sets of smaller subsystems, while maintaining inter-subsystem graph rigidity. 

Each subsystem is indexed by $l$ and we denote the covariance of the subsystem by $P^{(l)}$ and the state of the subsystem by $x^{(l)}$. Figure \ref{fig:subsys_ex} shows an example of how a network can be broken down into three subsystems. Subsystem $l=0$, which corresponds to node $n_0$, monitors only nodes $n_0$, $n_1$, and $n_2$ so that the subsystem state vector and covariance matrix will be of reduced dimension only containing information about estimates for those three nodes.  Similarly, subsystem $l=3$ monitors only nodes $n_1$, $n_3$, and $n_4$, whereas subsystem $l=4$ considers only nodes $n_3$ and $n_4$. The size of the state vectors and the covariance matrices are thus of a reduced form representative of the nodes defined to be in a given subsystem as shown in Figure \ref{fig:Pl}. 

The important step in this algorithm is to decompose the whole systems into smaller subsystems and let every node monitors only the nodes that are in its own subsystem. For the local error covariance matrices, it is known that $(P^{(l)})^{-1}$, which is the normal inverse of the reduced $P^l$ matrix, does not equal $(P^{-1})^{(l)}$ which is the submatrix of $P^{-1}$ and this represents a challenge in decomposing the whole system into subsystems. We will remove the superscript $l$ for simplicity from now.

The \algtwo{} algorithm addresses the previous challenge of distributed matrix inversions by employing the L-Banded inverse on the global covariance matrix, followed by the DICI-OR method presented in \cite{conf:dislarge} in order to approximate the inverse of the large covariance $P$ matrix. These two methods can be summarized as follows.

\begin{itemize}
\item \textbf{L-Banded Inverse}: We use the L-banded inversion provided in \cite{conf:dishermit} in order to obtain an L-banded approximate matrix $P^{-1^{k}}$ from   $P^k$. The L-banded structure is necessary in the application of the DICI-OR method stated below. The L-banded inverse method contains properties for approximating gaussian error process, which help with inverses of submatrices. This has also been presented in  \cite{conf:distest} and \cite{conf:dislband}.

\item \textbf{DICI-OR Method}: The DICI-OR method uses the L-banded structure of the $P{^{-1}}^k$ matrix in achieving distributed matrix inversion from $P{^{-1}}^k$ to $P^k$. With the assumption that the non L-band elements are not specified in the covariance matrix $P^k$, a collapse step is effected in order to fill in the non L-band elements. The iterate step is then employed to produce the elements that lie within the L-band of $P^k$. The iterate step can be repeated to achieve quicker convergence in the state estimates.
\end{itemize}

The \algtwo{} algorithm leverages L-banded inverse and the DICI-OR method to achieve distributed matrix inversion. Unlike \algone{} where global covariance matrices $P$ and state vectors $x$ are needed to be maintained at each node, these two methods enable us to decompose the problem into smaller parts. While \cite{conf:dislarge} proposes operations in the information filter domain, the algorithm has been adapted to run in the Kalman filter domain in this work. $M$ is defined as $diag(P_{i|i}^{{-1}^k})$ and $0\leq\gamma\leq1$ is the overrelaxation parameter in the initial conditions. Using equations \eqref{eqn:moura4} and \eqref{eqn:moura5} as initial conditions, we seek to move from $P^{{-1}^k}_{i|i-1}$ to $P^{k}_{i|i-1}$ using the DICI-OR method. This uses the collapse step to calculate non L-band elements before proceeding to the iterate step to reconstruct the local $P^k_{i|i-1}$ matrix. The subscripts $a$ and $b$ are representative of the row and column indices of the matrix. The DICI-OR method can be summarized as following:\\

\textbf{Collapse Step}: 
\begin{eqnarray}
p_{ab} &= &p_{a,b-1}\cdot p^{-1}_{a+1,b-1}\cdot p_{a+1,b}\label{eq:moura6}
\end{eqnarray}

\textbf{Iterate Step}: 
\begin{eqnarray}
|a-b|\leq L\nonumber\\
p_{ab,t+1} &= &\left\{\begin{matrix}
\mathbf{s}_{a}\mathbf{\tilde{p}}^{b}_{t} & a\neq b \\ 
\mathbf{s}_{a}\mathbf{\tilde{p}}^{b}_{t}+\gamma m^{-1}_{aa} & a=b  
\end{matrix}\right.
\end{eqnarray}
where $\mathbf{s}_{a}$ is the $a$th row of matrix S which is defined in the initial conditions of DICI-OR, i.e, equation \ref{eqn:moura4}. $\mathbf{\tilde{p}}^{b}$ is the $b$th row of matrix P, and $m_{aa}$ is the $a$th diagonal element of the diagonal matrix $M$. The iterations index is determined by $t$. This step is only done for elements within the L-band. The \algtwo{} can be divided mainly into three steps, namely, measurement update, publish measurements, and time update.


\begin{algorithm}[ht]
\begin{flushleft}
\textbf{Algorithm 2: \algtwo{}}\\
Start with $\hat{x}^k_{0|-1}=\mathbf{E}x_0$ and $P^k_{0|-1}= \Pi_0$ for all $k$, and compute at every node $k$:\\
\textbf{Initial Conditions for DICI-OR}:
\begin{eqnarray}
S_\gamma &= &(1-\gamma)I_{n\cdot n}\>+\>\gamma M^{-1}(M - P^{-1^{k}}_{i|i})\label{eqn:moura4}\\
P^k_{i|i} &= &P^k_{i-1|i-1}\label{eqn:moura5}
\end{eqnarray}
\textbf{Step 1}: Measurement update:
\begin{eqnarray}
P_{i|i-1}^{{-1}^k} &\xleftarrow{Lbandinv} &P_{i|i-1}^{k}\\
P_{i|i}^{{-1}^k} &= &P_{i|i-1}^{{-1}^k} + \sum_{j\in\mathcal{N}_{k}} H_{i}^{*^{k,j}}R_{i}^{{-1}^j}H_{i}^{k,j}\\
P_{i|i}^{^k}&\xleftarrow{DICIOR}&P_{i|i}^{{-1}^k}\\
\hat{x}_{i|i}^{k} &= &\hat{x}_{i|i-1}^{^k} + P_{i|i}^{^k}\sum_{j\in\mathcal{N}_{k}} H_{i}^{*^{k,j}}R_{i}^{{-1}^j}(y_{i}^{k,j} - h_{i}^{j}(\hat{x}_{i|i-1}^{^k}))\nonumber
\end{eqnarray}

\textbf{Step 2}: Publish measurements among neighbors which will replace their estimates.

\textbf{Step 3}: Time update:
\begin{eqnarray}
P_{i+1|i}^{k} &= &\bar{F}_{i}(\hat{x}_{i|i}^{k})P_{i|i}^{k}\bar{F}_{i}(\hat{x}_{i|i}^{k})^{*} + Q_{i}\\
\hat{x}^{k}_{i+1|i} &= &\bar{F}_{i}(\hat{x}_{i|i}^{k})\cdot \hat{x}^{k}_{i|i} + \bar{u}_{i}(\hat{x}^{k}_{i|i})
\end{eqnarray}
\end{flushleft}
\end{algorithm}


\subsection{\algthree} \label{sec:alg3}

\algthree{} operates by letting each node monitors its neighbor nodes only, so the size of $\hat{x}$ is limited locally by the number of connecting nodes, an improvement from the \algone{} algorithm. In this algorithm, the unconstrained minimizations are solved numerically and simultaneously at each node by computing the values for the parameters that satisfy the first order optimality conditions. \algthree{} can be summarize in the steps presented in Algorithm 3. A more detailed discussion with the theoretical results for convergence of the current algorithm can be found in \cite{JacobiCDC17}.

\begin{algorithm}
\begin{flushleft}
\textbf{Algorithm 3: \algthree{}}\\
Start with $\hat{x}^k_{0}=\mathbf{E}x_0$. At every time instant $i$, do at every node $k$:\\
\textbf{Step 1}: Get from neighboring nodes $j \in \mathcal{N}_k$ their estimated state $\hat{x}^j_i=\left[ \hat{\bm{p}}^{j_i^T},\: \hat{o}^j_i,\: \hat{b}^j_i\right]^{T}$.\\

\textbf{Step 2}: Find optimal estimate $\hat{x}^k_{i+1}$ of node $k$ based on the received estimates and measured distances:\\

\begin{align}
\left[\hat{o}^k_{i+1}~\hat{\bm{p}}^k_{i+1} \right] = \argminB_{o^k_i,\bm{p}^k_i} &\sum_{j \in \mathcal{N}_k} (d^{k,j}_{i}-( o^k_{i} - \hat{o}^j_{i} ) - \left\Vert \bm{p}^k_{i}-\hat{\bm{p}}^j_{i}\right\Vert _{2}/c)^2
\label{eq:opt1}\\
\left[\hat{o}^k_{i+1}~\hat{\bm{p}}^k_{i+1} \right] =\argminB_{o^k_i,\bm{p}^k_i} &\sum_{j \in \mathcal{N}_k} (t_{2}(i)-t_{3}(i)-( o^k_{i} - \hat{o}^j_{i} ) - \left\Vert \bm{p}^k_{i}-\hat{\bm{p}}^j_{i}\right\Vert _{2}/c)^2
\label{eq:opt2}\\
\left[\hat{b}^k_{i+1}~\hat{\bm{p}}^k_{i+1} \right] =\argminB_{b^k_i,\bm{p}^k_i} &\sum_{j \in \mathcal{N}_k} ( R^{k,j}_{i} - \left\Vert \bm{p}^k_{i}-\hat{\bm{p}}^j_{i}\right\Vert _{2} - c\cdot \tilde{R}^{k,j}_i )^2
\label{eq:opt3}
\end{align}
\textbf{Step 3}: Combine the results from the optimization step and send the new estimate $\hat{x}^k_{i+1}$ to the neighbors $j \in \mathcal{N}_k$.
\end{flushleft}
\end{algorithm}

The steps in Algorithm 3 have considered Type 3 messages which are more accurate. However, depending on the application considered, Type 2 can be enabled instead by replacing equation \ref{eq:opt3} with equation \ref{eq:opt4}. 

\begin{align}
\argminB_{b^k_i,\bm{p}^k_i} &\sum_{j=l} ( r^{j,k}-\left\Vert \bm{p}^k_{i}-\bm{p}^j_{i}\right\Vert _{2}-\frac{c}{2}\left(b^k_i-b^j_i \right)T_{RSP1} )
\label{eq:opt4}
\end{align}

\section{Evaluation} \label{sec:eval}

\begin{figure*}[tb]
\centering
\includegraphics*[width = 0.9\textwidth]{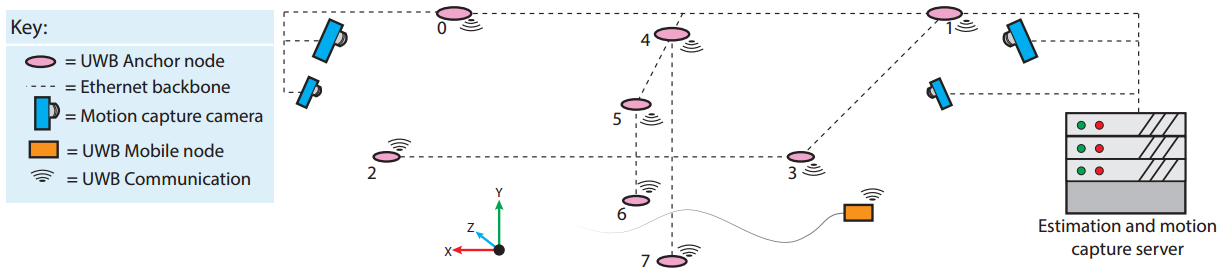}
\vspace{-0.3cm}
\caption{Experimental setup overview, including, UWB Anchor nodes, motion capture cameras, and mobile quadrotor UWB nodes}
\label{fig:full_sys}
\end{figure*}

\begin{figure*}
\centering
\begin{minipage}[t]{0.30\textwidth}%
\includegraphics[height=1.75in]{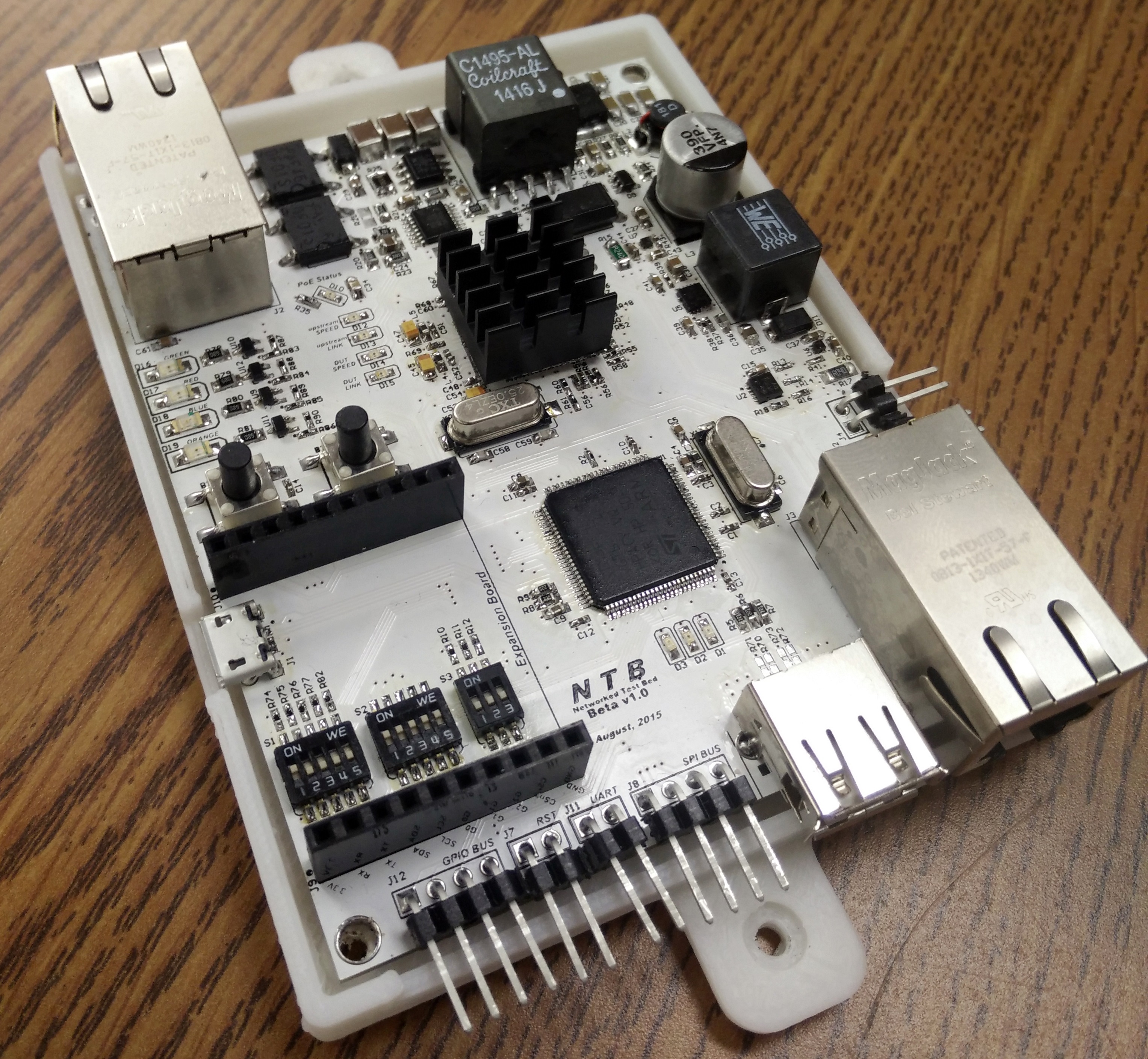}
\vspace{-0.3cm}
\caption{Custom anchor node with ARM Cortex M4 processor and UWB expansion.}
\label{fig:ntbanon}
\end{minipage}\hspace{5mm}
\begin{minipage}[t]{0.30\textwidth}%
\centering
\includegraphics[height=1.75in]{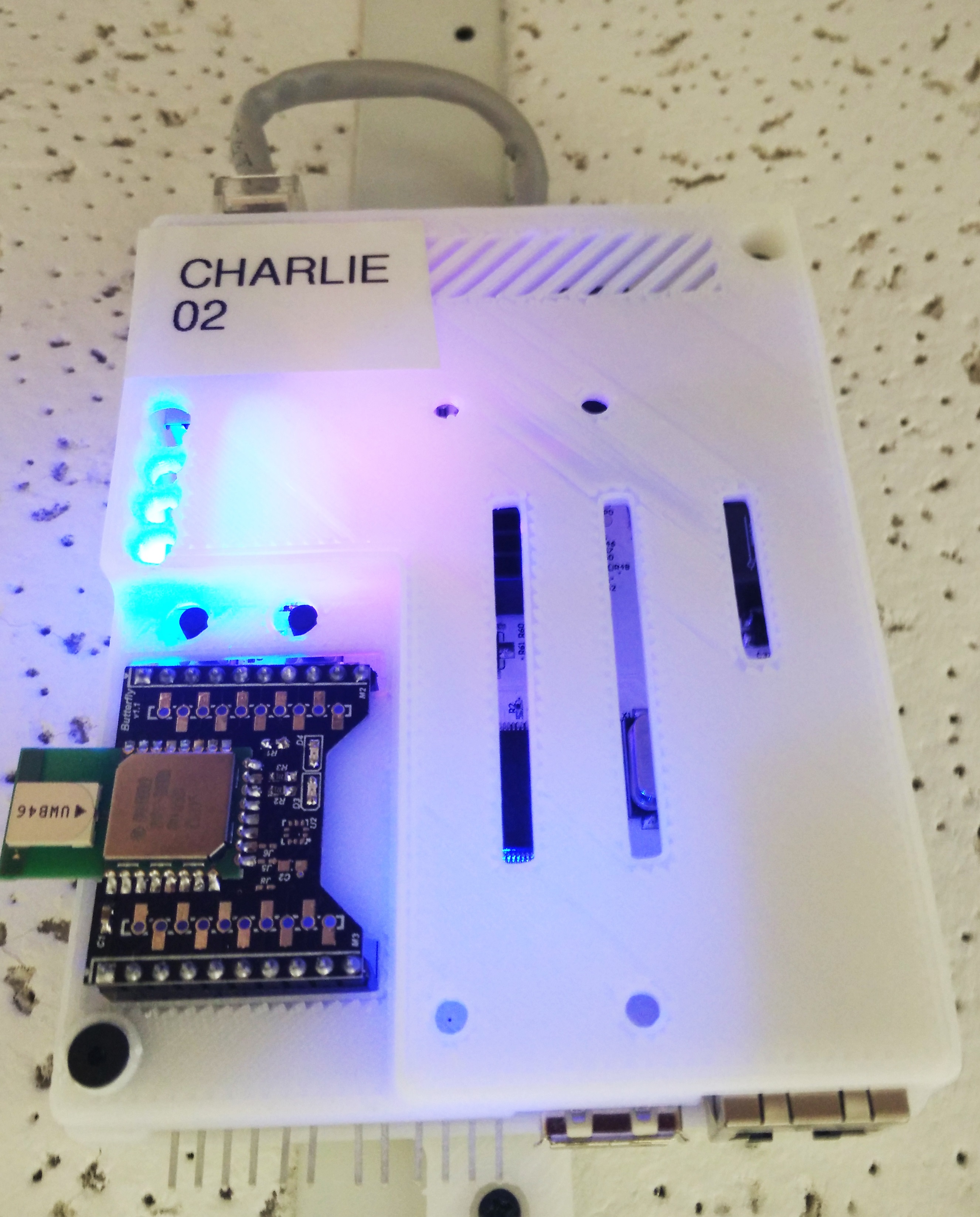}
\vspace{-0.3cm}
\caption{Ceiling-mounted anchor with DW1000 UWB radio in 3D-printed enclosure.}
\label{fig:ntbceil}
\end{minipage}\hspace{5mm}
\begin{minipage}[t]{0.30\textwidth}%
\centering
\includegraphics[height=1.75in]{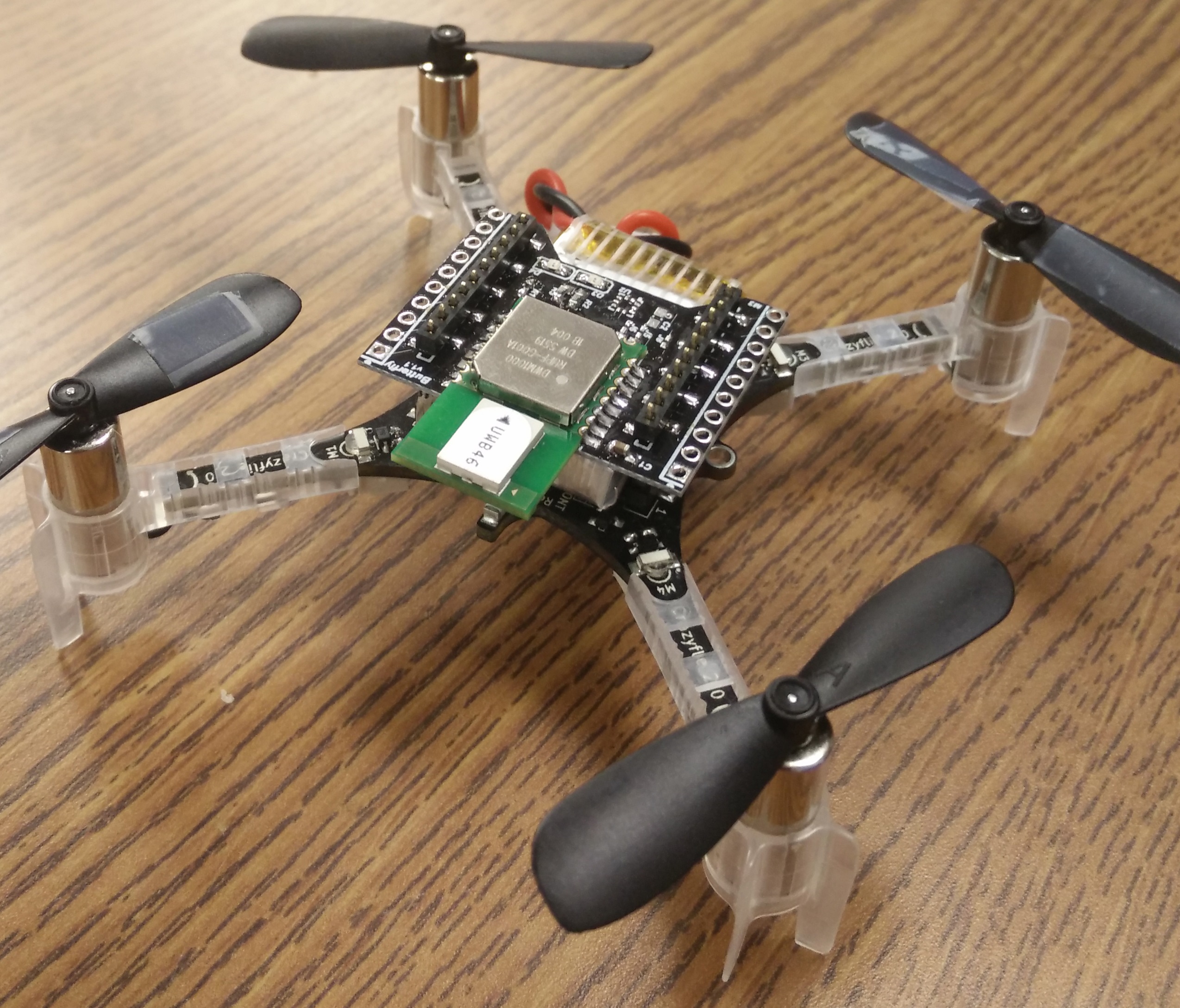}
\vspace{-0.3cm}
\caption{CrazyFlie 2.0 quadrotor helicopter with DW1000 UWB expansion.}
\label{fig:quad}
\end{minipage}
\end{figure*}

We evaluated the performance of the proposed algorithms in a custom ultra-wideband RF test bed based on the DecaWave DW1000 IR-UWB radio \cite{dw1000}. The setup is shown in Figure \ref{fig:full_sys}. We can summarize the main components of our test bed as follows: (i) Fixed \emph{anchor} nodes powered by an ARM Cortex M4 processor with Power over Ethernet (PoE) and an expansion slot for a custom daughter board containing the DW1000, as shown in Figure \ref{fig:ntbanon} and Figure \ref{fig:ntbceil}. This radio is equipped with a temperature-compensated crystal oscillator with frequency equals 38.4 MHz and a stated frequency stability of $\pm 2$ ppm. We installed eight UWB anchor nodes in different positions in a 10$\times$9 $m^2$ lab. Six anchors are placed on the ceiling (roughly 2.5 m high) and 2 were placed at waist height (about 1 m) to better disambiguate positions in the vertical axis. Each anchor node is connected to an Ethernet backbone both for power and for communication to the central server, and each is capable of sending messages of type 1, 2, and 3, and is fully controllable over a TCP/IP command structure from the central server. These nodes are placed so as to remain mostly free from obstructions, maximizing line-of-sight barring pedestrian interference. (ii) Battery-powered mobile nodes also with ARM Cortex M4 processors based on the CrazyFlie 2.0 helicopter \cite{crazyflie} and equipped with the very same DW1000 radio as shown in Figure \ref{fig:quad}; (iii) A motion capture system capable of 3D rigid body position measurement with less than 0.5 mm accuracy; and finally (iv) a centralized server for aggregation of UWB timing information and ground truth position estimates from the motion capture cameras. We adopt a right-handed coordinate system where $y$ is the vertical axis and $x$ and $z$ make up the horizontal plane. We used 2 Hz as a rate of broadcasting the local time of each node.

\subsection{Case Study: Static Nodes} \label{sec:static}

We  demonstrate \sys{} performance in distributed simultaneous localization and time synchronization of static nodes. To begin, nodes are placed in 8 distinct locations around the 10 $\times$ 9 $m^2$ area, roughly in the positions indicated by Figure \ref{fig:full_sys}. The goal of \sys{} is to accurately estimate the positions of all network devices relative to one another, as well as, the relative clock offsets and frequency biases. This relative localization, or graph realization as it is sometimes called, is a well-researched field. Local minima, high computational complexity, and restrictions on graph rigidity are the main challenges in graph realization problems \cite{grp_aspnes,grp_jackson}. \sys{} does not put restrictions on the connectivity of the graph. A centralized Kalman filter (CKAL) is implemented as a baseline for comparison.

\subsubsection{Position Estimation} 
The \sys{} algorithms find relative positions. Therefore, we first superimpose the estimated positions onto the true positions of each node by use of a Procrustes transformation \cite{procrustes}. Specifically, the network topology as a whole is rotated and translated without scaling, until it most closely matches the true node positions. Once transformed, the error of a given node's position is defined as the $\ell_2$ norm of the transformed position minus the true position.

\begin{figure*}[t]
    \centering
    \begin{tabular}{ p{0.3\textwidth}  p{0.3\textwidth}  p{0.3\textwidth}}
    	\resizebox{0.25\textwidth}{!}{
            \begin{subfigure}[h]{0.3\textwidth}
      \centering
        \includegraphics[scale=0.45]{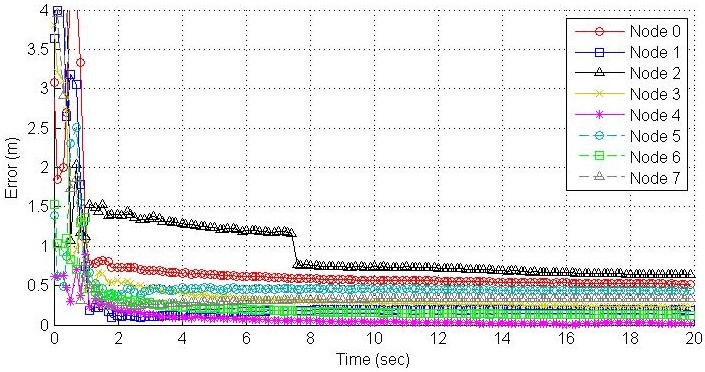}
        \caption{\algone}
    \end{subfigure}
	}
       &
	\resizebox{0.25\textwidth}{!}{
            \begin{subfigure}[h]{0.3\textwidth}
      \centering
        \includegraphics[scale=0.4]{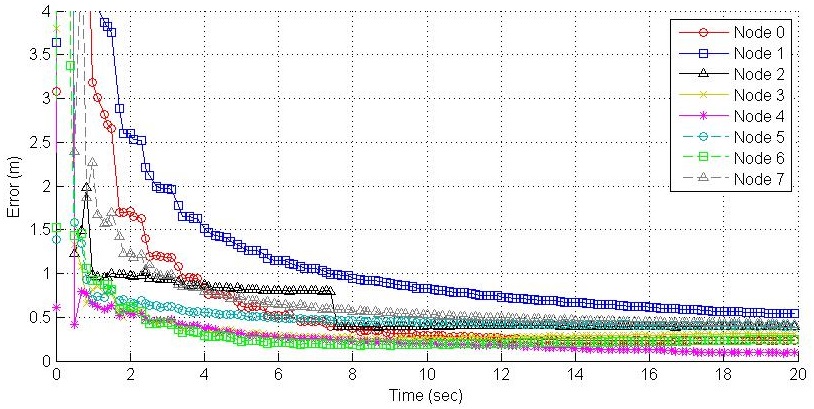}
        \caption{\algtwo}
    \end{subfigure}
  	 } 
   &
   \resizebox{0.25\textwidth}{!}{
            \begin{subfigure}[h]{0.3\textwidth}
      \centering
        \includegraphics[scale=0.45]{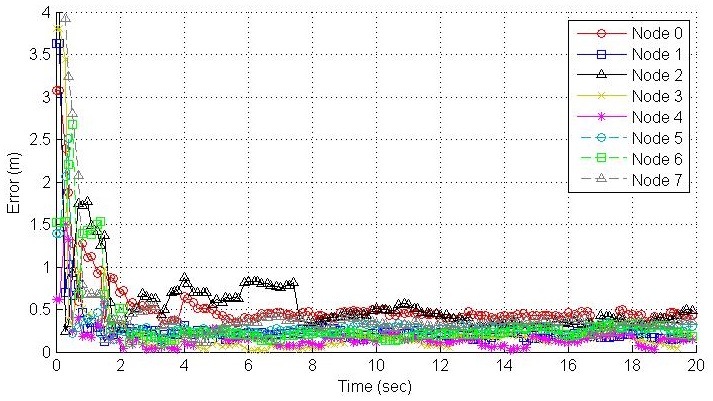}
        \caption{\algthree}
    \end{subfigure}
  	}
   \\ 
  \end{tabular}
  \vspace{-0.3cm}
\caption{Fully connected localization error.}
    \label{fig:fully}
\end{figure*}

We begin by showing the localization error of a fully connected network for all algorithms. Figure \ref{fig:fully} shows the localization errors for \algone{}, \algtwo{}, and \algthree{}, respectively, with Type 3 enabled for all algorithms. Enabling Type 3 inherits enabling Type 1 and Type 2, as Type 3 is a replicate of Type 1 and Type 2. Table \ref{tab:loc_err} summarizes the localization error of the eight nodes using \algone{}, \algtwo{}, \algthree{}, and CKAL. \algone{} achieves 0.311 $m$. While \algtwo{} and \algthree{} report 0.330 $m$ and 0.299 $m$, respectively. Also, we should not that enabling Type 2 only instead of Type 3 in our experiments does not have great effect of the overall performance.

\begin{figure}[h!]
\centering
\includegraphics[width = 0.4\textwidth]{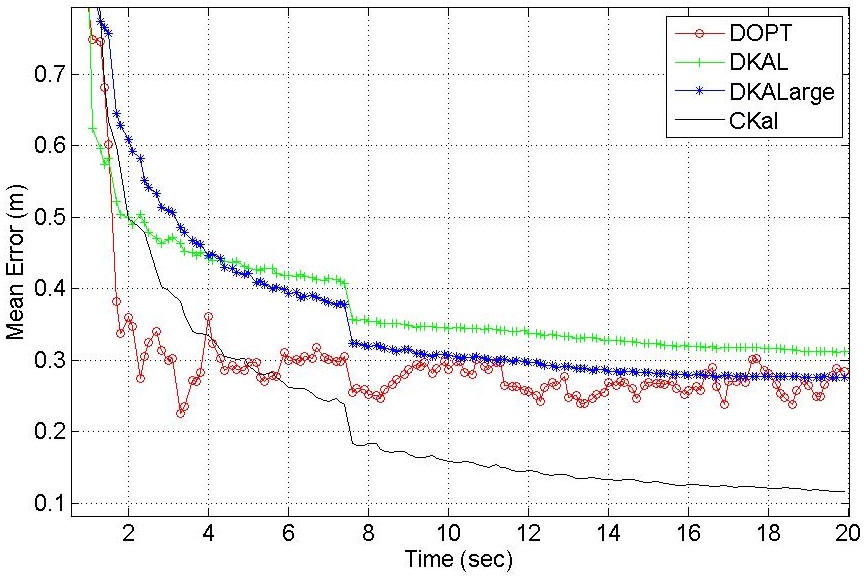}
\vspace{-0.3cm}
\caption{Average localization error with fully connected network.}
\label{fig:fully_connected_avg}
\end{figure}

\begin{figure}[h!]
\centering
\includegraphics[width = 0.4\textwidth]{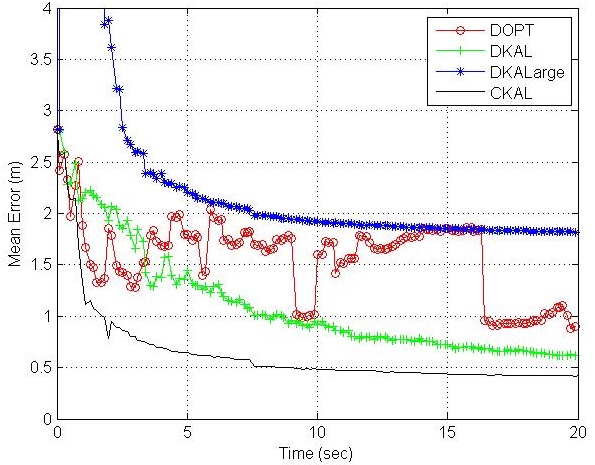}
\vspace{-0.3cm}
\caption{Average localization error where each node has 4 neighbors only.}
\label{fig:loc_err_4neig}
\end{figure}

We summarize the mean error of the position estimation for all nodes and for different localization algorithms in Figure \ref{fig:fully_connected_avg}. The centralized algorithm outperforms the distributed algorithms in \sys{}, as expected. Note that \algthree{} has the fastest convergence time, and \algone{} has the worst performance among the algorithms studied. In a second experiment, we show that \sys{} supports different network topologies without requiring the graph to be fully connected. Figure \ref{fig:loc_err_4neig} shows the localization error for the network topology where every node is only connected to four neighbors. \algtwo{} gave the worst performance as the DICI-OR algorithms is approximating the full covariance matrix inverse, as described before.

\begin{table*}[tbp]
\centering
\normalsize
\begin{tabular}{l||c|c|c|c|c|c|c|c||c|c}
 
Algorithm &node 0& node 1 & node 2& node 3&node 4&node 5&node 6& node 7& mean & std \\
\hline\hline
\algone 	&0.518&0.189 & 0.638 &0.228&0.021&0.433&0.126&0.336 &0.311&0.209\\
\algtwo 	&0.232&0.536&0.394&0.318&0.093&0.400&0.246&0.418 &0.330&0.137\\
\algthree   &0.402&0.202&0.530&0.193&0.156&0.309&0.199&0.397&0.299&0.133\\
CKAL 		&0.205&0.189&0.208&0.147&0.218&0.075&0.168&0.143&0.169&0.047\\
\end{tabular}
\caption{localization error (m) of different static nodes.}
\label{tab:loc_err}
\end{table*}

\subsubsection{Time synchronization}

\begin{figure}[h!]
\centering
\includegraphics[width = 0.44\textwidth]{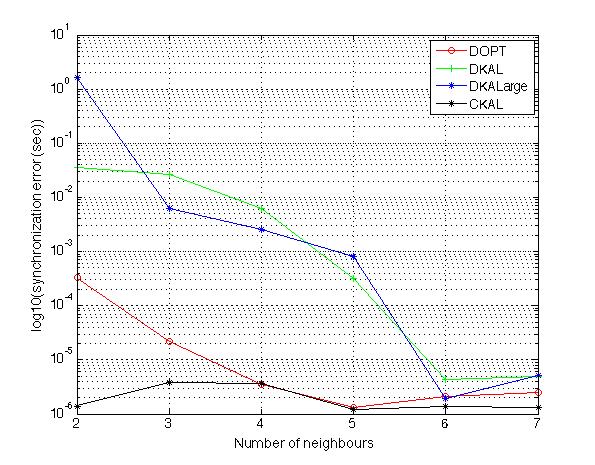}
\vspace{-0.3cm}
\caption{Log scale of the effect of decreasing the connectivity on the synchronization error.}
\label{fig:time_sync_neig_log}
\end{figure}

In order to test the time synchronization, we could configure a third party node to send a query message and compare the timestamp upon receiving that message as done in \cite{FTSP}. However, in order to decrease the uncertainty in our test mechanism, we choose the root node to do this job. This testing mechanism is better than others as reported in \cite{conf:uncer}, and been used also in \cite{conf:pulsesync}. As mentioned before, we choose node 0 to be the reference node, i.e, $b^0_{i}:=0$ and $o^0_{i}:=0~~\forall i$. Table \ref{tab:sync_err} shows the synchronization errors for all nodes with respect to node 0. \algone{} comes with the best performance, then \algtwo{} and \algthree{}. We should note that the synchronization errors in Table \ref{tab:sync_err} are reported for a fully connected network. The effect of decreasing the connectivity on the \sys{} algorithms is shown in Figure \ref{fig:time_sync_neig_log}. \algtwo{} starts in a good shape with a fully connected network, but then the effect of approximating the covariance inverse appears when decreasing the connectivity. \algthree{} performs better than \algone{} for smaller connectivity networks.
\vspace{-0.5cm}
\begin{table*}[tbp]
\centering
\normalsize
\begin{tabular}{l||c|c|c|c|c|c|c||c|c}
 
Algorithm & node 1 & node 2& node 3&node 4&node 5&node 6& node 7& mean & std \\
\hline\hline
\algone 	&0.807& 9.088& 1.868 &2.332& 5.936&9.624&5.342&5.000 &  3.502 \\
\algtwo 	&5.223& 6.448& 5.203 &5.339& 5.863& 3.313 &4.222&5.087 & 1.036 \\
\algthree   &2.15& 0.891&2.090&2.343&1.651&5.215&3.293&2.520 &1.391\\
CKAL 		&1.362&2.045  &1.440 &1.517 &1.792 &0.267&0.708 & 1.304  &  0.617\\
\end{tabular}
\caption{Synchronization error ($\mu$ seconds) of different nodes with respect to node 0.}
\label{tab:sync_err}
\end{table*}


\subsection{Case Study: Mobile Nodes} \label{sec:mobile}

We have shown that \sys{} can be used to localize and synchronize a network of static nodes. We now present the case of a heterogeneous network containing both static
and mobile nodes. To the eight anchor nodes topology used in Section \ref{sec:static}, we add one mobile node in the form of a CrazyFlie quadrotor as shown in Figure \ref{fig:quad}. We analyzed the results of running \sys{} based on Type 3 measurements with a fully connected graph.

\begin{figure}[h!]
\centering
\includegraphics[width = 0.51\textwidth]{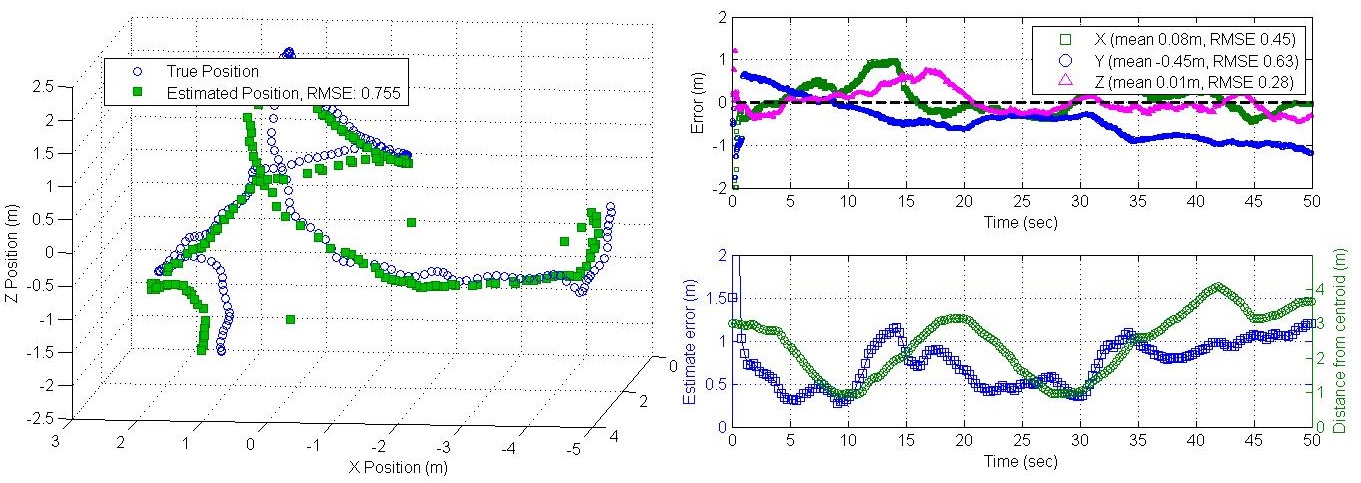}
\caption{Localization errors for \algone{} in 3D for a single mobile node. Spatial errors (left) are shown with corresponding per-axis errors by time (top right). Additionally, the error is plotted against the mobile node’s distance from the network centroid (bottom right).}
\label{fig:mobile_dekf}
\end{figure}

\begin{figure}[h!]
\centering
\includegraphics[width = 0.51\textwidth]{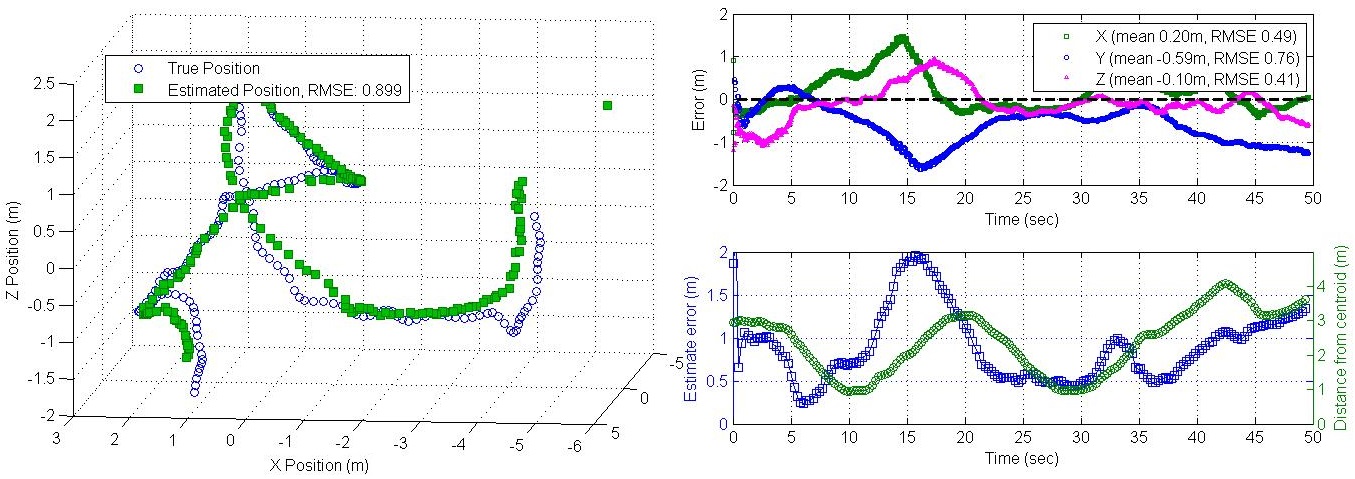}
\caption{Localization errors for \algtwo{} in 3D for a single mobile node. Spatial errors (left) are shown with corresponding per-axis errors by time (top right). Additionally, the error is plotted against the mobile node’s distance from the network centroid (bottom right).}
\label{fig:mobile_large}
\end{figure}

\begin{figure}[h!]
\centering
\includegraphics[width = 0.51\textwidth]{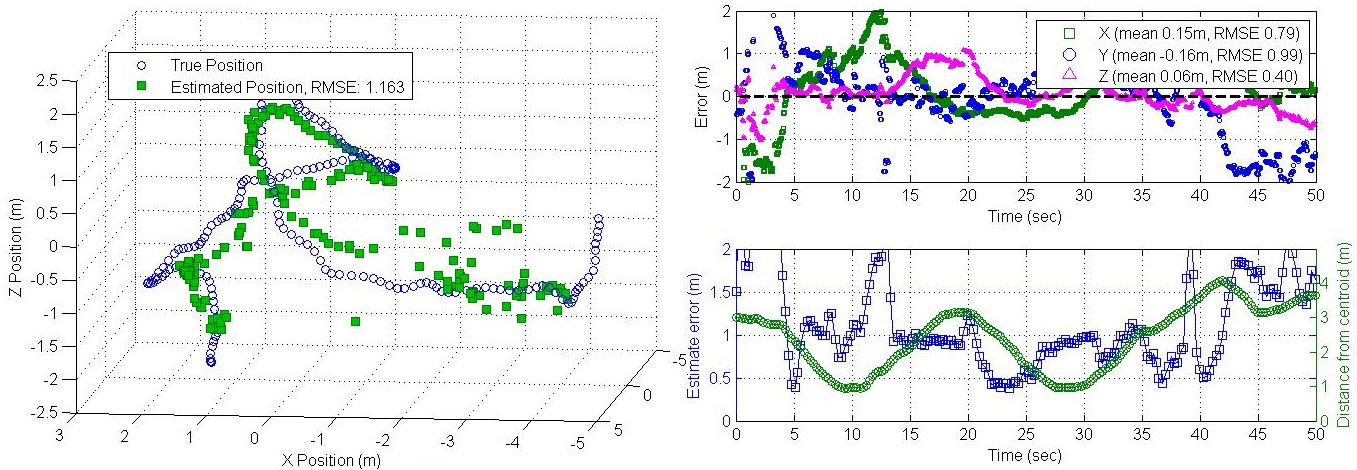}
\caption{Localization errors for \algthree{} in 3D for a single mobile node. Spatial errors (left) are shown with corresponding per-axis errors by time (top right). Additionally, the error is plotted against the mobile node’s distance from the network centroid (bottom right).}
\label{fig:mobile_est}
\end{figure}

A number of experiments were performed with quadrotors traveling with variable velocities. Figures \ref{fig:mobile_dekf}, \ref{fig:mobile_large}, and \ref{fig:mobile_est} show the results of traveling quadrotors self-localizing themselves for 2 minutes using \algone{}, \algtwo{}, and \algthree{}, respectively, using Type 3 messages. The left plot of the three figures show a 3D comparison of the self-localizing estimated position and the ground truth position reported by the motion capture cameras. \algone{} achieved the best self localization estimation with an RMSE of 75 cm. \algtwo{} and \algthree{} reported 89 cm and 116 cm, respectively. The top left subfigures of the three figures show the 3D localization error. The localization errors in each axis separately are shown in the top right sub figure. Finally, the location of the mobile node in the network determines to some extent the accuracy with which it can be localized. A device whose location is more central to the network (i.e. closer to the centroid defined as the mean of all node positions) is more likely to be fully constrained in terms of its relative position. This correlation can be loosely seen in the bottom right of the three figures.

\vspace{-3mm}



\section{Conclusion} \label{sec:conc}

This paper has described and evaluated \sys{}, an architecture for distributed simultaneous time synchronization and localization of static and mobile nodes in a networks. Three different algorithms were proposed, namely \algone{}, \algtwo{}, and \algthree{}, that perform distributed estimation in a scalable fashion. Several experiments using real, custom ultra-wideband wireless anchor nodes and mobile quadrotor nodes were conducted and they indicate that the proposed architecture is reliable in terms of performance, and efficient in the use of computational resources. \sys{} is made possible by state-of-the-art advances in commercial ultra-wideband radios, and continued improvements to these devices will further underscore the importance of treating temporal and spatial variables in a joint fashion. Future directions will deal with testing over real large scale network by considering more nodes. Also, secure estimation of nodes under malicious attacks is another direction.

\section*{Acknowledgments}

This research is funded in part by the National Science Foundation under awards CNS-1329755 and CNS-1329644. The U.S. Government is authorized to reproduce and distribute reprints for Governmental purposes notwithstanding any copyright notation thereon. The views and conclusions contained herein are those of the authors and should not be interpreted as necessarily representing the official policies or endorsements, either expressed or implied, of NSF, or the U.S. Government.




\bibliographystyle{ACM-Reference-Format}
{\small
\bibliography{ref}
}
%


\end{document}